\documentclass[10pt,twocolumn,letterpaper]{article}

\usepackage{iccv}
\usepackage{times}
\usepackage{epsfig}
\usepackage{graphicx}
\usepackage{amsmath}
\usepackage{amssymb}

\usepackage{makecell}
\usepackage{algorithm}
\usepackage{algpseudocodex}

\usepackage{subfigure}

\usepackage[numbers,sort&compress]{natbib}

\usepackage[pagebackref,breaklinks,colorlinks,bookmarks=false]{hyperref}

\usepackage[capitalize]{cleveref}
\crefname{section}{Sec.}{Secs.}
\Crefname{section}{Section}{Sections}
\Crefname{table}{Table}{Tables}
\crefname{table}{Tab.}{Tabs.}

\usepackage{multirow}
\usepackage{booktabs}

\iccvfinalcopy 

\def\iccvPaperID{11493} 

\ificcvfinal\fi

\begin{document}

\title{Strip-MLP: Efficient Token Interaction for Vision MLP}

\author{Guiping Cao$^{1,2}$  \quad  Shengda Luo$^1$ \quad
Wenjian Huang$^1$ \quad
Xiangyuan Lan$^{2,*}$ \quad \\
Dongmei Jiang$^2$ \quad
Yaowei Wang$^2$ \quad
Jianguo Zhang$^{1,2,}$\thanks{Corresponding author.}\\
$^1$Southern University of Science and Technology, Shenzhen, China\\
$^2$Peng Cheng Laboratory, Shenzhen, China\\
{\tt\small 12131099@mail.sustech.edu.cn, \{luosd, huangwj, zhangjg\}@sustech.edu.cn,}\\
{\tt\small \{lanxy, jiangdm, wangyw\}@pcl.ac.cn}
}

\maketitle
\ificcvfinal\thispagestyle{empty}\fi

\begin{abstract}
   Token interaction operation is one of the core modules in MLP-based models to
   exchange and aggregate information between different spatial locations. However, the power of token interaction on the spatial dimension is highly dependent on the spatial resolution of the feature maps, which limits the model's expressive ability, especially in deep layers where the feature are down-sampled to a small spatial size.
  To address this issue, we present a novel method called \textbf{Strip-MLP} to enrich the token interaction power in three ways. Firstly, we introduce a new MLP paradigm called Strip MLP layer that allows the token to interact with other tokens in a cross-strip manner, enabling the tokens in a row (or column) to contribute to the information aggregations in adjacent
  but different strips of rows (or columns).
  Secondly, a \textbf{C}ascade \textbf{G}roup \textbf{S}trip \textbf{M}ixing \textbf{M}odule (CGSMM) is proposed to overcome the performance degradation caused by small spatial feature size.
  The module allows tokens to interact more effectively in the manners of within-patch and cross-patch, which is independent to the feature spatial size. Finally, based on the Strip MLP layer, we propose a novel \textbf{L}ocal \textbf{S}trip \textbf{M}ixing \textbf{M}odule (LSMM) to boost the token interaction power in the local region.
  Extensive experiments demonstrate that Strip-MLP significantly improves the performance of MLP-based models on small
  datasets and obtains comparable or even better results on ImageNet. 
  In particular, Strip-MLP models achieve higher average Top-1 accuracy than existing MLP-based models by +2.44\% on Caltech-101 and +2.16\% on CIFAR-100.
  The source codes will be available at~\href{https://github.com/Med-Process/Strip_MLP}{https://github.com/Med-Process/Strip\_MLP}.

\end{abstract}

\vspace{-3ex}
\section{Introduction}
\label{sec:introduction}

\begin{figure}[h]
   \centering
    \includegraphics[width=0.70\linewidth]{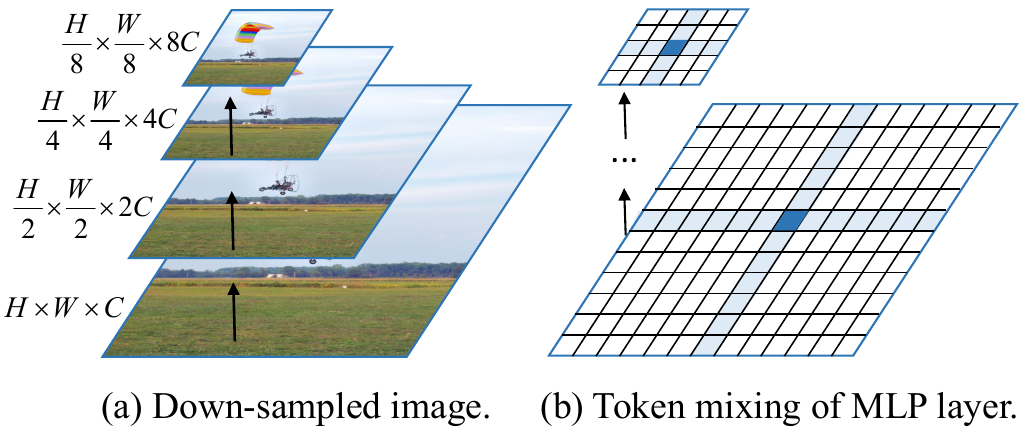}
    \caption{Token interaction of the MLP layer on the down-sampled image feature. (a) The process of which the feature resolution will be down-sampled from $H \times W \times C$ into a small spatial size of $\frac{H}{8} \times \frac{W}{8} \times 8C$. (b) Token interaction on different feature resolutions.
    }
    \label{fig:resolution}
    \vspace{-2ex}
 \end{figure}

\vspace{-2ex}
In computer vision, Convolutional Neural Networks (CNNs) are one of the most popular network backbones, which have made a series of breakthroughs~\cite{he2016deep}. 
Inspired by the great success of self-attention-based architectures~\cite{vaswani2017attention} in natural language processing (NLP), the Transformer models~\cite{dosovitskiy2020image, liu2021swin, wang2021pyramid} are introduced into the field of computer vision, and have achieved comparable results with state-of-the-art (SOTA) CNNs. 
Although ViT~\cite{dosovitskiy2020image} and its variants outperform traditional CNNs, the models introduce high computational complexity to construct attention maps.
Recently, some studies~\cite{liu2022convnet, tang2022sparse} in the vision community suggest that the attention mechanism is not necessary and simpler model architectures are proposed.

MLP-based models, like MLP-Mixer~\cite{tolstikhin2021mlp}, gMLP~\cite{liu2021pay}
and ViP~\cite{hou2022vision} process the data with the Multilayer Perceptrons (MLP),
showing great potential to improve the performance of vision models~\cite{touvron2022resmlp}.
As the first visual deep MLP network, MLP-Mixer~\cite{tolstikhin2021mlp} introduces two types of MLP layers: \emph{Channel-Mixing MLPs} (CMM) and \emph{Token-Mixing MLPs} (TMM). 
For CMM, the module mainly mixes the information between different channels of each token.
For TMM, it allows each spatial token to interact with all other tokens (the whole image) in a single MLP layer. 
However, this design also introduces a much larger number of parameters and higher computational complexity prone to overfitting. To address this problem, Sparse MLP (SMLP)~\cite{tang2022sparse} and Vision Permutator (ViP)~\cite{hou2022vision} propose a similar layer of parallel structure, 
which applies one dimension MLP along the axial directions, and parameters are shared among rows or columns, respectively. Therefore, it reduces the number of model parameters and computational complexity, avoiding the common over-fitting problem. 

Although SMLP and ViP alleviate some deficiencies of MLP-Mixer~\cite{tolstikhin2021mlp}, both methods bring the challenge that the token interaction power is highly dependent on the feature spatial size when interacting tokens on spatial rows (or columns). As shown in \cref{fig:resolution}, the spatial feature resolution is down-sampled to a small size but with more channels, which means the feature pattern of each token is mainly concentrated on the channel dimension rather than the spatial one. Interacting tokens along the spatial dimension by sharing the weights among all channels would seriously \emph{ignore the feature pattern differences among different channels, which may degrade the token interaction power}, especially in deep layers with small spatial feature resolution. Here, we mark this problem as \emph{the Token's interaction dilemma}.
Taking SMLP as an example, we analyze the feature resolution and complexity of the model in different stages in detail (seen in ~\cref{sec:parameter_complexity}). We find that as the spatial feature size decreases by down-sampling stage by stage, the token interaction layer also becomes smaller and smaller, which makes the token interaction power degraded rapidly.

To address the aforementioned challenges, in this paper, we propose a new efficient Strip MLP model, dubbed \textbf{Strip-MLP}, to enrich the power of the token interaction layer in three ways. For the level of a single MLP layer, inspired by the \emph{cross-block} normalization schemes of HOG~\cite{dalal2005histograms}, we design a Strip MLP layer to allow the token to interact with other tokens in a cross-strip manner, enabling each row or column of the tokens to contribute differently to other rows or columns.
 For the token interaction module level, we develop channel-wise group mixing of CGSMM,
 enabling the tokens in a row (or column) to contribute to the information aggregations in adjacent but different strips of rows (or columns).
 to tackle the problem that the token interaction power decreases in deep layers with the spatial feature size significantly reduced but with multiplying channels.
Considering the existing methods~\cite{tolstikhin2021mlp, tang2022sparse, hou2022vision} interact the tokens mainly in the long range of row (or column), which may not aggregate tokens well in the local region, we propose a new \textbf{L}ocal \textbf{S}trip \textbf{M}ixing \textbf{M}odule (LSMM) with a small Strip MLP unit to strengthen the token interaction power on local interactions.

The proposed Strip-MLP model significantly boosts the token interaction power, and the main contributions are:

\begin{itemize}
\vspace{-2ex}
\setlength{\itemsep}{0pt}
\setlength{\parsep}{0pt}
\setlength{\parskip}{0pt}

  \item[$\bullet$] A new MLP paradigm for vision MLP: Strip MLP layer, which aggregates
  the adjacent tokens in a cross-strip manner and enables each row or column of the tokens to contribute differently to other rows or columns, interacting tokens more efficiently.
  
  \item[$\bullet$] Designing a Cascade Group Strip Mixing module and a Local Strip Mixing module, which effectively improves the model's token interaction power and boosts the tokens aggregation in the local region, respectively;

  \item[$\bullet$] Extensive experiments show that Strip-MLP remarkably improves the performances of the MLP-based models. Strip-MLP achieves higher average Top-1 accuracy by +2.44\% on Caltech-101 and +2.16\% on CIFAR-100 over the existing MLP-based models. In addition, our models achieve comparable or even better performances on ImageNet-1K compared with traditional MLP-based models, and other popular CNNs and transformer-based models.
  
\vspace{-1ex}
\end{itemize}

\begin{figure*}[t]
   \centering
    \includegraphics[width=0.83\linewidth]{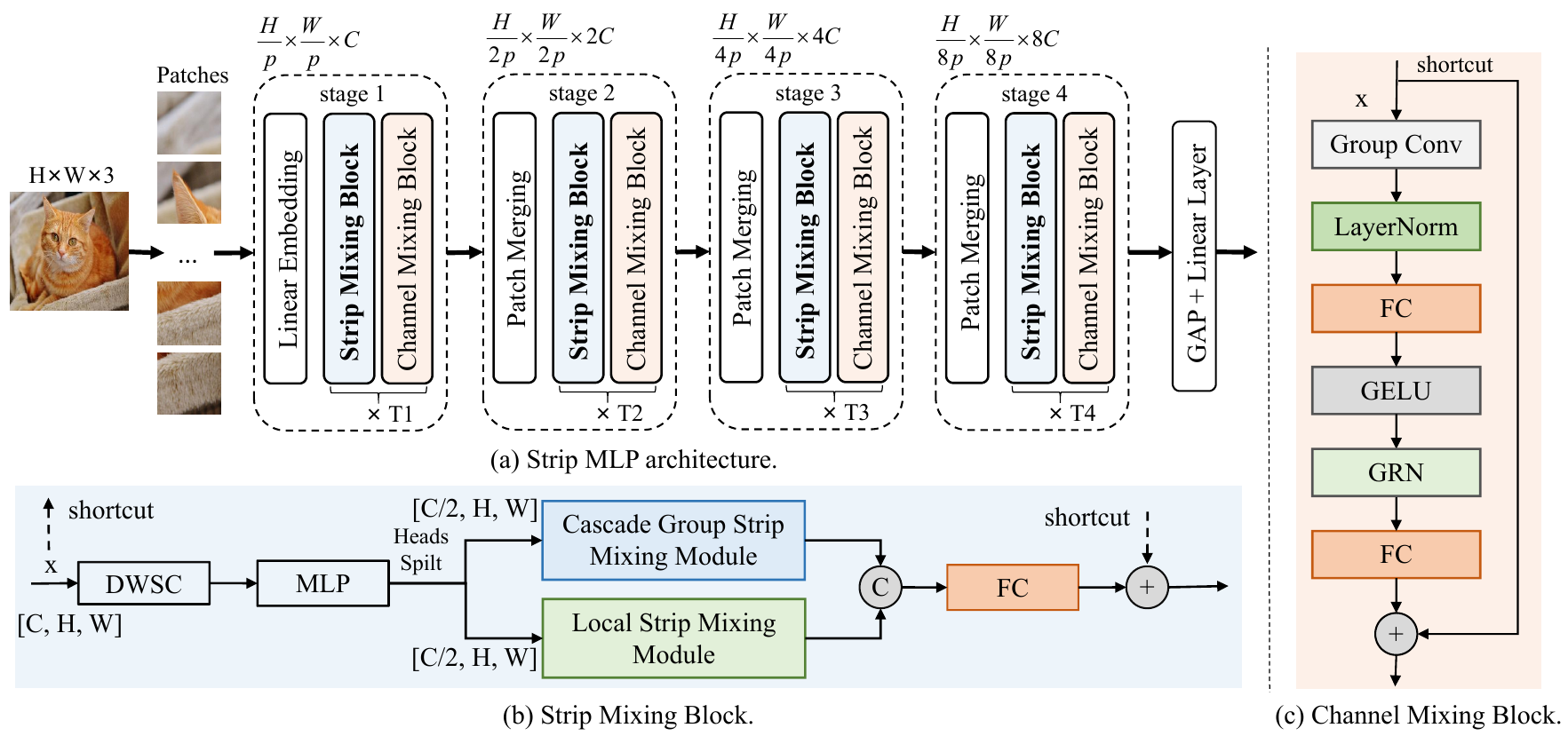}
    \caption{The overall and components architecture of Strip-MLP. (a) Strip-MLP has four stages, and $T1$ to $T4$ means the repeated times of the block in each stage. (b) The Strip Mixing Block splits the head (channel dimension) into two heads, and they are fed into the two parallel branches. (c) Channel Mixing Block architecture. 
    }
    \label{fig:strip_mlp}
    \vspace{-2ex}
 \end{figure*}

\section{Related Work}

\vspace{-1ex}
Deep neural networks for vision recognition can be mainly divided into three categories: Convolutional Neural Networks (CNNs),
Vision Transformers (ViTs), and Multi-Layer Perceptron-based models (MLPs)~\cite{hou2022vision}.

\vspace{0.1ex}

\textbf{CNNs-Based Models.} CNNs are the de-facto standard deep learning network models for 
vision tasks and have been intensely studied in the vision community. AlexNet~\cite{krizhevsky2017imagenet} is a symbolically significant model that won the ILSVRC 2012 contest with far higher performance than others. Since then, CNNs-based models have attracted more attentions, and lots of deeper and more effective architectures~\cite{he2016deep, howard2017mobilenets, simonyan2014very, szegedy2015going, tan2019efficientnet, wang2020deep, zhang2018shufflenet,zhang2023decoding} have been proposed. 
With the convolution and pooling layer, CNNs aggregate the feature
in a local region but not well in the long-term dependencies, which are optimized by the new vision model like Transformer~\cite{vaswani2017attention}
and MLP~\cite{tolstikhin2021mlp} models.

\vspace{0.1ex}

\textbf{Transformer-Based Models.} Transformer~\cite{vaswani2017attention} is introduced for machine translation and becomes the
reference model for all-natural language processing (NLP) tasks. Inspired by the great success of Transformer in NLP, 
ViTs~\cite{dosovitskiy2020image} first applies a standard Transformer to images, which attains excellent results 
compared to the SOTA CNNs model. 
DeiT~\cite{touvron2021training} 
introduces several training strategies and distillation methods to make ViTs more effective on the smaller ImageNet-1K dataset. 
By proposing a hierarchical Transformer with shifted windows, Swin Transformer~\cite{liu2021swin} achieves SOTA accuracy on ImageNet-1K, 
bringing greater efficiency by self-attention in local window and cross-window connection. 
For these models, self-attention is the core module but with heavy computational burdens to obtain an attention map.

\vspace{0.1ex}

\textbf{MLP-Based Models.} Without the convolutions and self-attention mechanism, MLP-Mixer~\cite{tolstikhin2021mlp} builds the architecture that only uses the MLP layer and achieves competitive performance on image classification benchmarks.
Since then, the researchers have developed many MLP-like variants~\cite{chen2021cyclemlp, guo2022hire, liu2021pay, tang2022sparse, tang2022image, touvron2022resmlp, 
tu2022maxim, wang2022dynamixer, zhang2021morphmlp} models. The work~\cite{liu2022we} gives a comprehensive survey about visual deep MLP models and compares the intrinsic connections and differences between convolution, self-attention mechanism, and Token-mixing MLP in detail.
Sparse MLP~\cite{tang2022sparse} introduced a sparse operation to separately
aggregate information along axial directions, avoiding the quadratic computational complexity of conventional MLP. Hire-MLP~\cite{guo2022hire}
presents a novel variant of MLP-based architecture via hierarchically rearranging tokens to aggregate local and global spatial information. 
Wave-MLP~\cite{tang2022image} represents each token as a wave function with two parts of amplitude and phase, which has the ability to model varying contents from different input images.
Above all, there is still the limitation of token interaction power degraded significantly especially when the spatial feature resolution becomes small, which has been overlooked by previous studies. In this paper, we aim to enrich the token interaction power on both the single MLP
layer and token interaction module to advance the performance of MLP-based models.


\section{Method}

\vspace{-1ex}
In this section, we first present the overall architecture of Strip-MLP. Then, we show the key proposed components: Strip MLP Layer, \textbf{C}ascade \textbf{G}roup \textbf{S}trip \textbf{M}ixing \textbf{M}odule (CGSMM) and \textbf{L}ocal \textbf{S}trip \textbf{M}ixing \textbf{M}odule (LSMM) of the model in detail, 
and present the comparison analysis between the Strip MLP and traditional MLP model in terms of parameters and complexity. Finally, we define four kinds of architecture variants to compare the performance of the models in different sizes.

\begin{figure*}[h]
   \centering
    \includegraphics[width=0.83\linewidth]{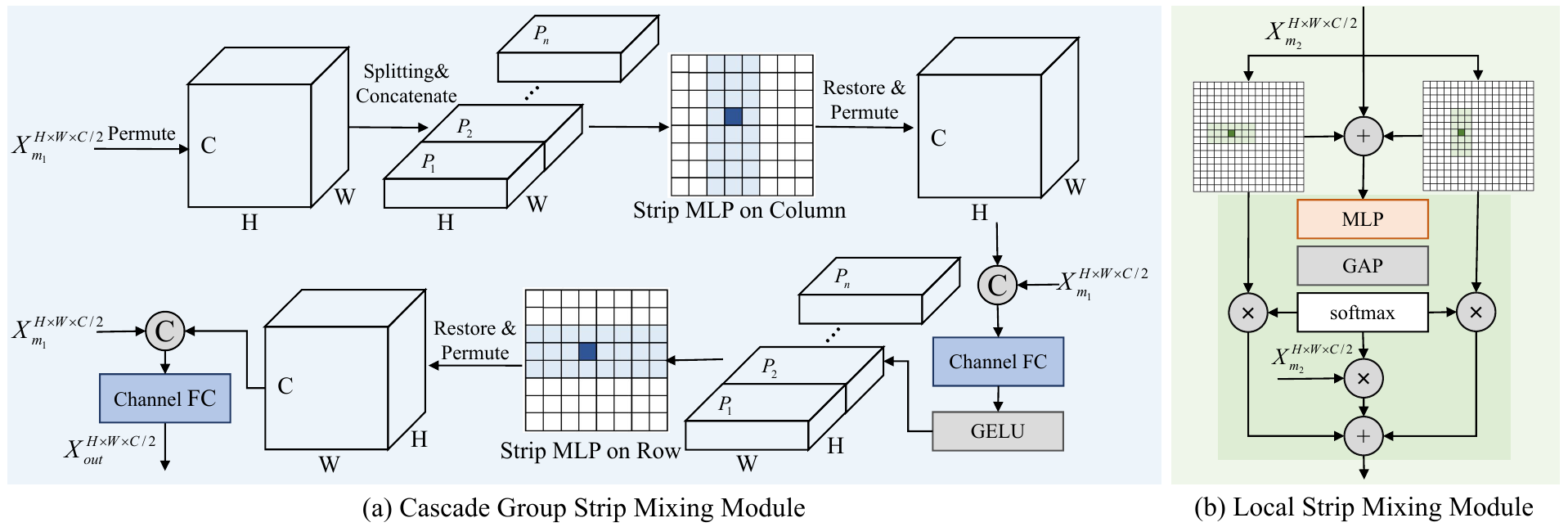}
    \caption{ (a) The architecture of CGSMM. It contains the Strip MLP layer, the feature splitting, restoring, and permuting operation in a cascade mode. (b) The structure of the Local Strip Mixing module. The dark green block composes of a re-weight module.
    }
    \label{fig:cascade_strip_mixing}
 \end{figure*}

\vspace{-1ex}
\subsection{Overall Architecture}
\label{sec:architecture}

\vspace{-1ex}
An overview of the Strip-MLP model is depicted in \cref{fig:strip_mlp} (a). 
We design the Strip-MLP as a hierarchical structure of a multi-stages model. Given the input image $I$, Strip-MLP model makes patch embedding and models the image feature in four stages.

\textbf{Patch Embedding.} Strip-MLP firstly splits the input image $I \in \mathbb{R}^{H \times W \times{3}}$ (H, W: image height and width) into a sequence of image patches (also referred as tokens) $I_p \in \mathbb{R}^{hw\times{c}}$ (c: the number of channels), where the patch size is $p \times p$ and patch number is $hw$ $(h = \frac{H}{p}, w = \frac{W}{p})$, and then all patches are linearly projected into a desired higher dimension $C$ of the feature ($X \in \mathbb{R}^{hw\times{C}}$).

\textbf{Mixing Block.} The purpose of mixing block is to boost the interactions between the features of diverse spatial locations and channels. To effectively aggregate spatial and channel information, we design two sub-blocks of the \emph{Strip Mixing Block} and \emph{Channel Mixing Block}. The Strip Mixing Block is comprised of CGSMM and LSMM, which blend and aggregate spatial information more efficiently at the global and local level, respectively.
Patch merging module aims to merge the feature in which the spatial dimension will be reduced by $2 \times 2$, and the channel dimension increases by 2 times so that the model down-samples the feature from $\frac{H}{p} \times \frac{W}{p}$ into $\frac{H}{8p} \times \frac{W}{8p}$ stage by stage. To obtain multi-scale features, we apply a single convolutional layer to the output features of stages 1 and 2. Then, we add the resulting features to the input features of stages 3 and 4.

\textbf{Head Layer.} The features extracted by the multiple blocks are fed into a global average pooling (GAP) layer to reduce the feature dimension. Finally, the feature will be fed into a fully-connected head layer for classification.


\vspace{-1ex}
\subsection{Strip Mixing Block and Channel Mixing Block}

\vspace{-1ex}
\textbf{Strip Mixing Block.} To improve the token interaction power, we design the block to aggregate both long and short range interactions in a parallel manner. As illustrated in \cref{fig:strip_mlp} (b), we split the feature in channel dimension, with one-half of the channel feature being fed into CGSMM to model the long-range interactions, and the remaining feature being fed into LSMM to aggregate the local interactions. Given the input feature $X \in \mathbb{R}^{H \times W \times C}$, the block can be formulated as:

\vspace{-2ex}
\begin{equation}
   X_m = MLP((DWSC(X)))
   \label{eq:eq1}
 \end{equation}
 
\vspace{-5ex}
\begin{equation}
\hspace{-1ex}Y=FC(Cat(CGSMM(X_{m_1}^{\frac{C}{2}}), \!LSMM(X_{m_2}^{\frac{C}{2}}))) \!+ \!X
\vspace{-2ex}
\label{eq:eq2}
\end{equation}
where $X_m$ and $Y$ are the intermediate and output feature of the block. $X_m$ is split into $X_{m_1}$ and $X_{m_2}$ with half channel, respectively. DWSC means depth-wise convolution~\cite{chollet2017xception}, and the kernel size is $3\times3$. MLP is a serial connection of
fully-connection (FC), batch normalization~\cite{ioffe2015batch} and GELU~\cite{hendrycks2016gaussian} activation layers. $Cat(*)$ denotes concatenation operation.

\textbf{Channel Mixing Block.} The block aims to aggregate the information between channels of the token, and the basic structure is shown in \cref{fig:strip_mlp} (c). Following the inverted bottleneck in~\cite{liu2022convnet} and global response normalization (GRN) in~\cite{woo2023convnext}, we design the Channel Mixing block to increase the contrast and selectivity of channels.
\vspace{-1ex}

\subsection{CGSMM and LSMM}

\vspace{-1ex}
We first introduce the Strip MLP Layer.
Based on this layer, we design CGSMM using a simple but effective strategy that splits the feature into patches along channel dimension and interacts tokens more effectively in the manners of within-patch and cross-patch, which is independent to the feature spatial size.
In addition, the design of existing methods ~\cite{tolstikhin2021mlp, tang2022sparse, hou2022vision} only allows tokens to interact in a long range of rows (or columns) with sharing weight; thus the resulting model may struggle with effectively aggregating both global and local information simultaneously. Therefore, 
we design LSMM to capture the local interactions more efficiently.

\textbf{Strip MLP Layer.} In MLP-based models, the majority of MLP layers treat each row and column of the data in isolation, as formulated in \cref{eq:eq3}, which may lead to the inefficiency of token interaction. Inspired by the \emph{cross-block} normalization schemes of HOG~\cite{dalal2005histograms}, which overlaps the blocks and enables each scalar cell response to contribute into different blocks, we propose the Strip MLP layer. The proposed layer applies MLP on the ``strip" data of adjacent rows or columns in the spatial direction in order to aggregate the feature in a cross-strip manner.
Given the input $X$, we formulated the Strip MLP layer (setting strip width as 3 for example) in \cref{eq:eq4,eq:eq5}:

\vspace{-3ex}
\begin{equation}
   X^{h}_{*,j} = W_1X_{*,j};    \quad X^{w}_{i, *} = W_2X_{i,*}
   \label{eq:eq3}
 \end{equation}

\vspace{-3.5ex}
\begin{equation}
   X^{h}_{*, j} = W_3Cat(X_{*,j-1}, X_{*,j}, X_{*,j+1})
   \label{eq:eq4}
 \end{equation}

\vspace{-3ex}
\begin{equation}
   X^{w}_{i, *} = W_4Cat(X_{i-1,*}, X_{i,*}, X_{i+1,*})
   \label{eq:eq5}
\vspace{-1ex}
 \end{equation}
where $W_1 \sim W_4$ are the weights of the MLP layer, $i$ and $j$ are the token index in row and column.
The superiorities of the Strip MLP layer are mainly in two aspects: on one hand, Strip MLP layer enables the token to interact with other tokens in both short and long spatial ranges simultaneously. 
On the other hand, similar to the HOG cross-block normalization process, each row (or column) not only serves for the current row (or column) tokens aggregation but also makes contribution to the adjacent rows (or columns) feature aggregation. For example, in \cref{eq:eq4}, $X_{*,j}$ makes different contributions for aggregating the processed feature of $X^{h}_{*, j-1}$, $X^{h}_{*, j}$ and $X^{h}_{*, j+1}$, making tokens interacting more efficiently in a cross-strip manner.

\textbf{Cascade Group Strip Mixing Module.} The architecture is shown in \cref{fig:cascade_strip_mixing} (a). The module applies the Strip MLP layer in row and column direction in a cascade mode. As the operation of the Strip MLP layer on row is similar to that on column, we take one of them to show the method.

\emph{Patch Splitting of the Data along Channel Dimension.} Given the input feature $X_{m_1} \in \mathbb{R}^{H \times W \times \frac{C}{2}}$, the module firstly permutes the feature into $X^{H \times \frac{C}{2} \times W}_{m_1}$, and splits the feature into $P$ patches in the channel dimension and concatenated along the column ($X^{H \times \frac{C}{2P} \times PW}$) dimension.

\emph{Group Strip MLP Layer.} To improve the token interaction power, we propose a \emph{Group Strip MLP Layer} (GSML) to interact tokens in the manners of within-patch and cross-patch. In particular, we apply the unshared weights of Strip MLP on different patches, and the tokens within the same patch share the weights for interaction. Then, we restore the feature to original shape and concatenate it with the input feature. To interact the tokens crossing patches, we apply a channel fully-connected (Channel FC) layer to interact tokens between different patches.

\textbf{Local Strip Mixing Module.} The module is illustrated in \cref{fig:cascade_strip_mixing} (b). In order to better aggregate the local interactions on spatial dimension, we define a small Strip MLP unit, where the strip width and length are $3$ and $7$, respectively.
Given the input feature $X_{m_2}^{H \times W \times \frac{C}{2}}$, we aggregate the local interactions in rows and columns direction simultaneously. Following \cite{chen2021cyclemlp,hou2022vision, tang2022image}, we sum all branches with a re-weight module.


\subsection{Parameter and Complexity Analysis}
\label{sec:parameter_complexity}

\vspace{-1ex}
In multi-stage processing architecture, 
\emph{the Token's interaction dilemma} becomes more serious, as the feature spatial resolution will be down-sampled in deep layers, and the token interaction power will be weakened. In this section, we present the comparison analysis of parameters and complexity to show the effectiveness of the proposed CGSMM.
Considering the majority MLP-Based models have the similar model structure of token interaction, we analyze the parameters and complexities with Sparse MLP~\cite{tang2022sparse}, which is a popular model with good performance on various datasets.

Sparse MLP~\cite{tang2022sparse} firstly applies MLP on the columns and rows of $X$ to map $\mathbb{R}^{CW \times{H}}$ to $\mathbb{R}^{CW \times{H}}$ separately. Then, the model concatenates the two processed features with the input feature $X$ and uses a linear layer on the channel dimension to fuse the feature from $\mathbb{R}^{HW \times{3C}}$ to $\mathbb{R}^{HW \times{C}}$. The number of parameters and FLOPs of the first step are ${W}^2 + {H}^2$ and ${CHW}{(H + W)}$; they are  $3{C^2}$ and $3HW{C^2}$ for the fusion step.
Comparatively, the number of parameters and FLOPs of Strip MLP layer for token interaction are 
$3P(H^{2}+W^{2})$ 
and 
$3CHW(H+W)$, respectively. The fusion step has
$4C^{2}$ parameters and  
$4HWC^{2}$ FLOPs.
In particular, for $H_1=W_1=56, H_4=W_4=7, C_1=112, C_4=112\times{2^3}=896,$ and $P=\frac{C}{4}$, we calculate the number of parameters and FLOPs of both two layers. As shown in \cref{tab:parameters_analysis}, for the Sparse MLP block, the number of parameters for token interaction in stage 4 (only $0.10k$) decreases by 62.70 times than in stage 1.
In addition, most of the number of the block's parameters and the FLOPs concentrate in the fusion step.
For example, only 0.01\% parameters (0.52\% FLOPs) are used for the token interaction step in stage $4$.
Based on the above analysis, to improve the token interaction power, it would be better to redesign the block to balance the number of parameters and FLOPs.

\begin{table}
  \centering
  \renewcommand\arraystretch{0.4}
  \setlength{\tabcolsep}{1.2mm}{
    \begin{tabular}{c | c | c | c | c }
    \toprule
    \multirow{2}{*}{Items} & \multicolumn{2}{c}{Sparse MLP} & \multicolumn{2}{c}{Strip MLP} \\
    ~ & Params & FLOPs & Params & FLOPs \\
    \midrule
    Stage 1 & 6.27$k$ & 39.34M & 526.85$k$ & 118.01M \\
    Stage 4 & 0.10$k$ & 0.62M & 65.86$k$ & 1.84M \\
    Changes & $\downarrow$62.70 & $\downarrow$63.45 & \textbf{$\downarrow$8.00} & $\downarrow$64.14 \\
    \midrule
    Stage 4 Fusion & 2.41M & 118.01M & 3.21M & 157.35M \\
    Proportion & 0.01\% & 0.52\% & \textbf{2.01\%} & \textbf{1.16\%} \\ 
    \bottomrule
  \end{tabular}}
  \caption{The model parameters and complexity comparison between Sparse MLP and Strip MLP layer in different stages.
  \emph{Stage 1} and \emph{Stage 4} show the parameters (Params) and floating-point operations (FLOPs) for token interaction in
  one layer between different stages. \emph{Stage 4 Fusion} indicates the Params and FLOPs in the feature fusion step of the token interaction module.
  \emph{Proportion} means the ratio between the token interaction step and other steps in the token interaction module of stage 4.}
  \vspace{-1ex}
  \label{tab:parameters_analysis}
\end{table}

In CGSMM, we use a simple but effective strategy that splits the feature along channel dimension into patches and interacts the tokens in the manners of within-patch and cross-patch. No matter how the spatial resolution decreases, the module can still interact the tokens with channel-wise specificity in different patches. In \cref{tab:parameters_analysis}, we improve the \emph{Proportion} of the number of parameters in the token interaction layer from 0.01\% to 2.01\% in stage 4.
Although our GSML brings more parameters and computations when just considering one token interaction layer, the total number of parameters and overall computational complexity of the model are decreased as the token interaction power is improved and we set small model configurations of $T_1 \sim \ T_4$.
Additionally, the experiments in \cref{sec:Experiments} show that our models achieve better performance with fewer parameters and FLOPs than other SOTA models.

\vspace{-1ex}
\subsection{Architecture Variants}

\vspace{-1ex}
We developed four variants of our Strip-MLP network: Strip-MLP-T$^{*}$ (light tiny), Strip-MLP-T (tiny), Strip-MLP-S (small), Strip-MLP-B (base), which have similar or even smaller model sizes, compared with MLP-baed models~\cite{guo2022hire, tang2022sparse, tang2022image} and Swin Transformer~\cite{liu2022swin}, respectively.
The hyper-parameters of the four variants of the models are as follows:
\begin{itemize}
\setlength{\itemsep}{0pt}
\setlength{\parsep}{0pt}
\setlength{\parskip}{0pt}
  \item[$\bullet$] Strip-MLP-T$^{*}$: $C = 80$,$\{T_1 \sim T_4 \} = \{2, 2, 6, 2 \}$;
  \item[$\bullet$] Strip-MLP-T: $C = 80$,$\{T_1 \sim T_4 \} = \{2, 2, 12, 2 \}$;
  \item[$\bullet$] Strip-MLP-S: $C = 96$,$\{T_1 \sim T_4 \} = \{2, 2, 18, 2 \}$;
  \item[$\bullet$] Strip-MLP-B: $C = 112$,$\{T_1 \sim T_4 \} = \{2, 2, 18, 2 \}$;
\vspace{-1ex}
\end{itemize}
where $C$ means the channel number of the hidden layers in the first stage, and $T_1 \sim T_4$ is the number of times repeated for Strip Mixing Block and Channel Mixing Block in each of the four stages, as shown in \cref{fig:strip_mlp}.

\vspace{-1ex}
\section{Experiments}
\label{sec:Experiments}

\vspace{-1ex}
We compare our Strip-MLP with three main categories of networks: CNNs-based, Transformer-based, and MLP-based models. We first show the experimental settings of the datasets and implementation details. Then, the proposed method is compared with other popular methods on three datasets. Finally, we ablate the critical design components of Strip-MLP network. 

\subsection{Experimental Settings}

\textbf{Datasets.} Methods are evaluated in three datasets which vary significantly in the number of training images:

\begin{itemize}
\vspace{-1ex}
\setlength{\itemsep}{0pt}
\setlength{\parsep}{0pt}
\setlength{\parskip}{0pt}

  \item[$\bullet$] Caltech-101~\cite{fei2006one}: it is a benchmark dataset that includes $101$ classes, totally around $9k$ images. We randomly take $80\%$ of each class of the data as the training set and use the remaining data as the testing set. 
  \item[$\bullet$] CIFAR-100~\cite{krizhevsky2009learning}: it has $100$ classes containing $600$ images of each class (totally $60k$), and there are $500$ training images and $100$ testing images per class.
  \item[$\bullet$] ImageNet-1K~\cite{deng2009imagenet}: it is an image classification benchmark containing $1.28M$ training images and $50k$ testing images
  of $1000$ classes.
\vspace{-1ex}
\end{itemize}

\begin{table}
   \centering
     \renewcommand\arraystretch{0.4}
         \setlength{\tabcolsep}{2mm}{
     \begin{tabular}{c | c | c | c }
     \toprule
     Method & Params & FLOPs & Top-1(\%) \\
     \midrule
     \multicolumn{4}{c}{CNNs-Based} \\
     \midrule
     ResNet50~\cite{he2016deep} & 23.71M & 4.12G & 89.61 \\
     ResNet101~\cite{he2016deep} & 42.71M & 7.85G & 88.56 \\
     ResNet152~\cite{he2016deep} & 58.35M & 11.58G & 89.04 \\
     \midrule
     \multicolumn{4}{c}{Transformer-Based} \\
     \midrule
     ViT-B/16~\cite{dosovitskiy2020image} & 85.85M & 16.86G & 53.96 \\
 
     Swin-T~\cite{liu2021swin} & 27.60M & 4.36G & 80.40 \\
     Swin-S~\cite{liu2021swin} & 48.91M & 8.52G & 79.83 \\
     Swin-B~\cite{liu2021swin} & 86.85M & 15.14G & 78.42 \\
     \midrule
     \multicolumn{4}{c}{MLP-Based} \\
     \midrule
     Wave-MLP-T~\cite{tang2022image} & 16.73M & 2.48G & 86.89 \\
     Hire-MLP-T~\cite{guo2022hire} & 17.58M & 2.16G & 87.85 \\
     Strip-MLP-T$^{*}$(\textbf{ours}) & 17.67M & 2.53G & \textbf{90.11} \\
     \midrule
     Wave-MLP-S~\cite{tang2022image} & 30.25M & 4.55G & 87.43 \\
     Hire-MLP-S~\cite{guo2022hire} & 32.65M & 4.26G & 88.48 \\
     Sparse-MLP-T~\cite{tang2022sparse} & 23.55M & 5.02G & 90.54 \\
     Strip-MLP-T(\textbf{ours}) & 24.33M & \textbf{3.67G} & \textbf{90.93} \\
     \midrule
     Wave-MLP-M~\cite{tang2022image} & 43.59M & 7.93G & 88.45 \\
     Hire-MLP-B~\cite{guo2022hire} & 57.77M & 8.17G & 88.62 \\
     Sparse-MLP-S~\cite{tang2022sparse} & 47.87M & 10.36G & 91.07 \\
     Strip-MLP-S(\textbf{ours}) & 43.66M & \textbf{6.83G} & \textbf{92.09} \\
     \midrule
     Wave-MLP-B~\cite{tang2022image} & 62.90M & 10.27G & 88.81 \\
     Hire-MLP-L~\cite{guo2022hire} & 95.03M & 13.50G & 88.19 \\
     Sparse-MLP-B~\cite{tang2022sparse} & 65.07M & 14.04G & 91.67 \\
     Strip-MLP-B(\textbf{ours}) & 58.49M & \textbf{9.22G} & \textbf{92.26} \\
     \bottomrule
   \end{tabular}}
   \caption{Image classification results of our Strip-MLP and other models on Caltech-101.
   \emph{Top-1} denotes the Top-1 accuracy. Here, the patch number of CGSMM is C/2.}
   \label{tab:classification_caltech101}
     \vspace{-1ex}
 \end{table}
 
 \textbf{Training Details.} In all of our experiments, the input image size is $224 \times 224$, and the input patch size is $4 \times 4$.
 Based on the deep-learning framework of PyTorch~\cite{paszke2019pytorch}, we train our models from scratch without any extra data for pre-training except for
 the transfer learning experiments. We employ an AdamW~\cite{kingma2014adam} optimizer with $300$ epochs for all training datasets and apply a cosine decay learning rate scheduler and $30$ epochs of linear warm-up.
 In addition, we adopt the same augmentation and regularization strategies as to Swin-Transformer~\cite{liu2021swin}.

 \subsection{Image Classification on Small Datasets}
 \vspace{-1ex}
 We conduct experiments on Small Datasets of Caltech-101 and CIFAR-100. The experiment results consistently indicate that our Strip-MLP model efficiently improves the token interaction power, and advances the MLP-based models' performance on the small dataset than other models.

\textbf{Results on Caltech-101.} \cref{tab:classification_caltech101} presents the image classification resluts of three types models on Caltech-101 dataset.
 Compared to ResNet~\cite{he2016deep}, all the four variants of our Strip-MLP models achieve higher Top-1 accuracy with fewer parameters and FLOPs, demonstrating an average increase of +2.69\% over three widely-used ResNet models.
 
 For the Transformer-based models, \eg Swin Transformer~\cite{liu2021swin}, Strip-MLP models noticeably surpass all variants models: +10.53\%/12.26\%/13.84\% over Swin-Transformer (T/S/B) models. ViT-B/16~\cite{dosovitskiy2020image}
 only gets an accuracy of 53.96\%, indicating that transformer-based models depend more on the scale of the datasets.

 When compared to the MLP-based models, like
 Sparse MLP~\cite{tang2022sparse}, Wave-MLP~\cite{tang2022image} and Hire-MLP~\cite{guo2022hire}, our models still achieve the best accuracy with fewer model parameters and FLOPs. For example, with fewer FLOPs (-4.28G), Strip-MLP-B (92.26\%) has an increase of +3.64\% Top-1 accuracy than Hire-MLP-B (88.62\%).

 \textbf{Results on CIFAR-100.} \cref{tab:classification_CIFAR100} compares the image classification results on CIFAR-100 with three kinds of architecture models.
 Compared with CNNs-based models~\cite{he2016deep}, Strip-MLP obtains average higher accuracy of +2.05\%$\sim$+2.43\%
 with a smaller number of parameters and FLOPs.
 When compared to the self-attention based model, \eg Swin Transformer~\cite{liu2021swin}, our model shows great superiority that
 Strip-MLP-B achieves 86.40\% Top-1 accuracy, +9.64\% than the Swin-B model with fewer parameters (58.49M vs. 86.85M) and lower FLOPs (9.22G vs. 15.14G). 
 In addition, our models keep the superiority over the other MLP-based models in that
 Strip-MLP achieves the best Top-1 accuracy (with a higher range of +1.04\%$\sim$+3.01\%) compared to the MLP-based models, \eg Wave-MLP~\cite{tang2022image}, Hire-MLP~\cite{guo2022hire}, and Sparse MLP~\cite{tang2022sparse},  with fewer parameters and FLOPs.
 
 \begin{table}
   \centering
     \renewcommand\arraystretch{0.5}
     \setlength{\tabcolsep}{2mm}{
     \begin{tabular}{c | c | c | c }
     \toprule
     Method & Params & FLOPs & Top-1(\%) \\
     \midrule
     \multicolumn{4}{c}{CNNs-Based} \\
     \midrule
     ResNet50~\cite{he2016deep} & 23.71M & 4.12G & 83.03 \\
     ResNet101~\cite{he2016deep} & 42.71M & 7.85G & 83.75 \\
     ResNet152~\cite{he2016deep} & 58.35M & 11.58G & 84.35 \\
     \midrule
     \multicolumn{4}{c}{Transformer-Based} \\
     \midrule
     Swin-T~\cite{liu2021swin} & 27.60M & 4.36G & 78.56 \\
     Swin-S~\cite{liu2021swin} & 48.91M & 8.52G & 78.10 \\
     Swin-B~\cite{liu2021swin} & 86.85M & 15.14G & 76.76 \\
     \midrule
     \multicolumn{4}{c}{MLP-Based} \\
     \midrule
     Wave-MLP-T~\cite{tang2022image} & 16.73M & 2.48G & 82.77 \\
     Hire-MLP-T~\cite{guo2022hire} & 17.58M & 2.16G & 82.44 \\
     Strip-MLP-T$^{*}$(\textbf{ours}) & 17.67M & 2.53G & \textbf{85.10} \\
     \midrule
     Wave-MLP-S~\cite{tang2022image} & 30.25M & 4.55G & 83.34 \\
     Hire-MLP-S~\cite{guo2022hire} & 32.65M & 4.26G & 82.94 \\
     Sparse-MLP-T~\cite{tang2022sparse} & 23.55M & 5.02G & 84.19 \\
     Strip-MLP-T(\textbf{ours}) & 24.33M & \textbf{3.67G} & \textbf{85.23} \\
     \midrule
     Wave-MLP-M~\cite{tang2022image} & 43.59M & 7.93G & 83.95 \\
     Hire-MLP-B~\cite{guo2022hire} & 57.77M & 8.17G & 83.17 \\
     Sparse-MLP-S~\cite{tang2022sparse} & 47.87M & 10.36G & 84.27 \\
     Strip-MLP-S(\textbf{ours}) & 43.66M & \textbf{6.83G} & \textbf{86.18} \\
     \midrule
     Wave-MLP-B~\cite{tang2022image} & 62.89M & 10.27G & 84.23 \\
     Hire-MLP-L~\cite{guo2022hire} & 95.03M & 13.50G & 83.53 \\
     Sparse-MLP-B~\cite{tang2022sparse} & 65.07M & 14.04G & 84.46 \\
     Strip-MLP-B(\textbf{ours}) & 58.49M & \textbf{9.22G} & \textbf{86.40} \\
     \bottomrule
   \end{tabular}}
   \caption{Image classification results of different models on CIFAR-100. The patch number of CGSMM is C/2.}
   \label{tab:classification_CIFAR100}
     \vspace{-3ex}
 \end{table}

 \subsection{Transfer Learning on Small Datasets}

 \vspace{-1ex}
 Following the transfer learning experiment of DynaMixer~\cite{wang2022dynamixer},
 we show the advantages of our models on the transfer learning tasks.
 With the pre-trained Strip-MLP-T$^{*}$ model on the ImageNet-1K, we transfer the model to downstream datasets of CIFAR-10 and CIFAR-100. Compared to the previous models, such as ViT~\cite{dosovitskiy2020image}, ViP~\cite{hou2022vision} and
 DynaMixer~\cite{wang2022dynamixer} models, our Strip-MLP-T$^{*}$ models achieve the best accuracies of 98.8\% and 89.4\% on two datasets, as shown in \cref{tab:classification_transfer_learning}, +1.7\%/0.6\% than ViT-S/16 and DynaMixer-S on CIFAR-10, +2.3\%/0.8\% over 
 ViT-S/16 and DynaMixer-S on CIFAR-100. More importantly, the number of parameters of our model is only 18M, which is far less than other models, indicating that our model is more efficient and effective.

 \begin{table}
   \centering
   \renewcommand\arraystretch{0.4}
    \setlength{\tabcolsep}{1.4mm}{
     \begin{tabular}{c | c | c | c }
     \toprule
     Model & Dataset & Params & Top-1(\%) \\
     \midrule
     ViT-S/16~\cite{dosovitskiy2020image} & \multirow{5}{*}{CIFAR-10} & 49M & 97.1 \\
     TST-14~\cite{yuan2021tokens} & ~ & 22M & 97.5 \\
     ViP-S/7~\cite{hou2022vision} & ~ & 25M & 98.0 \\
     DynaMixer-S~\cite{wang2022dynamixer} & ~ & 26M & 98.2 \\
     Strip-MLP-T$^{*}$(\textbf{ours}) & ~ & \textbf{18M} & \textbf{98.8} \\
     \midrule
     ViT-S/16~\cite{dosovitskiy2020image} & \multirow{5}{*}{CIFAR-100} & 49M & 87.1 \\
     TST-14~\cite{yuan2021tokens} & ~ & 22M & 88.4 \\
     ViP-S/7~\cite{hou2022vision} & ~ & 25M & 88.4 \\
     DynaMixer-S~\cite{wang2022dynamixer} & ~ & 26M & 88.6 \\
     Strip-MLP-T$^{*}$(\textbf{ours}) & ~ & \textbf{18M} & \textbf{89.4} \\
     \bottomrule
   \end{tabular}}
   \caption{The transfer learning results of the model pre-trained on ImageNet-1K and fine-tuned to CIFAR-10 and CIFAR-100.}
   \label{tab:classification_transfer_learning}
     \vspace{-2ex}
 \end{table}

 \begin{table}
   \centering
   \setlength{\tabcolsep}{0.75mm}{
     \renewcommand\arraystretch{0.4}
     \begin{tabular}{c | c | c | c | c }
     \toprule
     Method & Params & FLOPs & \makecell{Throughtput\\(image/s)} & \makecell{Top-1 \\ (\%)}  \\
     \midrule
     \multicolumn{5}{c}{CNNs-Based} \\
     \midrule
     ResNet50~\cite{he2016deep} & 26M & 4.1G & - & 78.5 \\
     ResNet101~\cite{he2016deep} & 45M & 7.9G & - & 79.8 \\
     RegNetY-4G~\cite{radosavovic2020designing} & 21M & 4.0G & 1157 & 80.0 \\
     RegNetY-8G~\cite{radosavovic2020designing} & 39M & 8.0G & 592 & 81.7 \\
     RegNetY-16G~\cite{radosavovic2020designing} & 84M & 16.0G & 335 & 82.9 \\
 
     \midrule
     \multicolumn{4}{c}{Transformer-Based} \\
     \midrule
     DeiT-S~\cite{touvron2021training} & 22M & 4.6G & 940 & 79.8 \\
     DeiT-B~\cite{touvron2021training} & 86M & 17.5G & 292 & 81.8 \\
     
     Swin-T~\cite{liu2021swin} & 29M & 4.5G & 755 & 81.3 \\
     Swin-S~\cite{liu2021swin} & 50M & 8.7G & 437 & 83.0 \\
     Swin-B~\cite{liu2021swin} & 88M & 15.4G & 278 & 83.5 \\
     \midrule
     \multicolumn{4}{c}{MLP-Based} \\
     \midrule
     ResMLP-S12~\cite{touvron2022resmlp} & 15M & 3.0G & 1415 & 76.6 \\
     gMLP-S~\cite{liu2021pay} & 20M & 4.5G & - & 79.6 \\
     CycleMLP-B1~\cite{chen2021cyclemlp} & 15M & 2.1G & 1038 & 78.9 \\
     Hire-MLP-Tiny~\cite{guo2022hire} & 18M & 2.1G & 1562 & 79.7 \\
     Wave-MLP-T~\cite{tang2022image} & 17M & 2.4G & 1208 & 80.6 \\
     Strip-MLP-T$^{*}$(\textbf{ours}) & 18M & 2.5G & 814 & \textbf{81.2} \\
 
     \midrule
     ResMLP-S24~\cite{touvron2022resmlp} & 30M & 6.0G & 715 & 79.4 \\
     Sparse-MLP-T~\cite{tang2022sparse} & 24M & 5.0G & 639 & 81.9 \\
     CycleMLP-B2~\cite{chen2021cyclemlp} & 27M & 4.0G & 641 & 81.6 \\
     Hire-MLP-Small~\cite{guo2022hire} & 33M & 4.2G & 808 & 82.1 \\
     Wave-MLP-S~\cite{tang2022image} & 30M & 4.5G & 720 & 82.6 \\
     Strip-MLP-T(\textbf{ours}) & 25M & \textbf{3.7G} & 597 & 82.2 \\
 
     \midrule
     CycleMLP-B4~\cite{chen2021cyclemlp} & 52M & 10.1G & 321  & 83.0 \\
     Sparse-MLP-S~\cite{tang2022sparse} & 49M & 10.3G & 361 & 83.1 \\
     Wave-MLP-M~\cite{tang2022image} & 44M & 7.9G & 413 & 83.4 \\
     Strip-MLP-S(\textbf{ours}) & 44M & \textbf{6.8G} & 381 & 83.3 \\
 
     \midrule
     ResMLP-B24~\cite{touvron2022resmlp} & 116M & 23.0G & 231 & 81.0 \\
     gMLP-B~\cite{liu2021pay} & 73M & 15.8G & - & 81.6 \\
     CycleMLP-B5~\cite{chen2021cyclemlp} & 76M & 12.3G & 247 & 83.2 \\
     Sparse-MLP-B~\cite{tang2022sparse} & 66M & 14.0G & 278 & 83.4 \\
     Hire-MLP-Base~\cite{guo2022hire} & 58M & 8.1G & 441 & 83.2 \\
     Wave-MLP-B~\cite{tang2022image} & 63M & 10.2G & 341 & 83.6 \\
     Strip-MLP-B(\textbf{ours}) & 57M & 9.2G & 300 &\textbf{83.6} \\
 
     \bottomrule
   \end{tabular}}
   \caption{Classification results of three kinds of models on ImageNet-1K without extra data.
   The group number of CGSMM is C/4. Throughput is tested on a single V100 GPU following \cite{liu2021swin}.}
   \label{tab:classification_ImageNet1k}
   \vspace{-3ex}
 \end{table}

 \subsection{Image Classification on ImageNet-1K}

 \vspace{-1ex}
 We conduct experiments on ImageNet-1K to show the effectiveness of the method.
 \cref{tab:classification_ImageNet1k} shows the performance together with the complexity (in terms of FLOPS and the number of parameters) of three types of models. 
 Strip-MLP outperforms the CNNs-based model: +0.5\% with half the number of parameters (25M vs. 39M) and FLOPs (3.7G vs. 8.0G) for Strip-MLP-T over RegNetY-8G~\cite{radosavovic2020designing}.
 Compared to Transformer-based models, \eg Swin Transformer~\cite{liu2021swin}, Strip-MLP achieves comparable results but with obvious superiority in terms of the number of parameters and FLOPs, such as Strip-MLP-B achieves similar performance to Swin-B but with fewer parameters (57M vs. 88M)
 and FLOPs (9.2G vs. 15.4G).
 Compared to the MLP-based models, Strip-MLP still shows the same efficiency in the various models.
 The Strip-MLP-T$^{*}$ achieves average higher accuracy by +2.12\% than other MLP-based models (81.2\% vs. 79.08\%). Both Strip-MLP-B and Wave-MLP-B~\cite{tang2022image} models get the same accuracy of 83.6\%. However, Strip-MLP-B uses fewer parameters (57M vs. 63M) and FLOPs (9.2G vs. 10.2G), indicating that the token-interaction power of our model is more efficient than other MLP-based models.
 
 In addition, we apply the Strip MLP layer into Wave-MLP-T (namely Wave-Strip-MLP-T) on ImageNet-1K, and we set the strip width as 3. The Wave-Strip-MLP-T gets better results than the original Wave-MLP-T model: 81.2\% vs. 80.6\%. The performance improvement by +0.6\% shows that our Strip MLP makes
 token interaction more efficient and can easily be used in other MLP models.

 \vspace*{-0.2cm}
 \subsection{Ablation Studies}

 \vspace{-1ex}
 To better show the effectiveness of the proposed method, we ablate the key components of the model design. Due to the limited GPU resources, we do ablation studies on the datasets of both Caltech-101 and CIFAR-100.
 
 \textbf{The effect of Strip Width in Strip MLP.} The strip width affects the token interaction ranges and determines the field of each row or column's contribution to adjacent tokens. We further conduct experiments to verify its effect by varying the width from 1 to 7 with a step at 2. \cref{tab:strip_width} shows the performances of our models on Caltech-101~\cite{fei2006one} and CIFAR-100~\cite{krizhevsky2009learning}, respectively.
 When the width becomes larger, the performance increases and tends to saturate, indicating that Strip MLP layer improves the token interaction power. 
 In all of our comparison experiments in previous sections, we consistently set the strip width as $3$ according to the ablation experiment results.
 
 \begin{table}
   \centering
     \setlength{\tabcolsep}{1.4mm}{
     \renewcommand\arraystretch{0.4}
     \begin{tabular}{ c | c | c | c | c }
     \toprule
     Strip Width & Dataset & Params & FLOPs & Top-1(\%) \\
     \midrule
     1 & \multirow{5}{*}{Caltech-101} & 41.33M & 6.71G & 91.38 \\
     3 & ~ & 43.66M & 6.83G & 92.09 \\
     5 & ~ & 45.99M & 6.96G & 91.72 \\
     7 & ~ & 48.33M & 7.08G & 91.86 \\
     \midrule
     1 & \multirow{5}{*}{CIFAR-100} & 40.71M & 6.71G & 86.09 \\
     3 & ~ & 41.88M & 6.83G & 86.50 \\
     5 & ~ & 43.04M & 6.96G & 86.39 \\
     7 & ~ & 44.21M & 7.08G & 86.35 \\
     \bottomrule
   \end{tabular}}
   \caption{Ablation study on the strip width of Strip MLP. We randomly select the model of Strip-MLP-S for Caltech-101 and Strip-MLP-S for CIFAR-100. }
   \label{tab:strip_width}
     \vspace{-2ex}
 \end{table}

 \textbf{Effects of Patch Number in CGSMM.} Different patch number brings varying degrees of token interaction power improvement and affects the model performance.
 In \cref{tab:classification_groups}, we design five decreased patch numbers from C into $1$ to show the effectiveness of CGSMM.
 Without the patch splitting ($P\!=\!1$) and GSML operation, the model performance decreases by 1.16\% on Caltech-101 (92.09\% $\rightarrow$ 90.93\%) and decreases by 1.47\% on CIFAR-100 (86.50\% $\rightarrow$ 85.03\%), which consistently proves the effectiveness of CGSMM in improving the token interaction power.
 From the experimental results, we can observe that the optimal patch number is different on the datasets with scale differences, so the optimal patch number should be determined through validation experiments. We consistently set the patch number as C/4 in other ablation studies.

 \begin{table}
   \centering
   \renewcommand\arraystretch{0.4}
       \setlength{\tabcolsep}{2mm}{
     \begin{tabular}{c | c | c | c | c }
     \toprule
     Dataset & Patches & Params & FLOPs & Top-1(\%) \\
     \midrule
     \multirow{3}{*}[-1.5ex]{Caltech-101} & C/1 & 47.22M & 6.83G & 91.81 \\
     & C/2 & 43.66M & 6.83G & 92.09 \\
     & C/4 & 41.88M & 6.83G & 91.38 \\
     & C/8 & 40.99M & 6.83G & 91.33 \\
     & 1 & 40.16M & 6.83G & 90.93 \\

     \midrule
     \multirow{3}{*}[-1.5ex]{CIFAR-100}  & C/1 & 47.22M & 6.83G & 86.20 \\
                                         & C/2 & 43.66M & 6.83G & 86.18 \\
                                         & C/4 & 41.88M & 6.83G & 86.50 \\
                                         & C/8 & 40.99M & 6.83G & 85.98 \\
                                         & 1 & 40.16M & 6.83G & 85.03 \\
     \bottomrule
   \end{tabular}}
   \caption{Ablation study about the patch number of the CGSMM. The base model is Strip-MLP-S. 
   $C/1$ means splitting the feature $C$ into C patches in channel dimension, namely $P=C/1=C$.}
   \label{tab:classification_groups}
     \vspace{-3ex}
 \end{table}

 \textbf{Cascade vs. Parallel Architecture of GSML.}  Applying the GSML in a cascade architecture enables the token interacts with other tokens across the entire 2D space in just one module, which needs two modules for the parallel structure so that may decrease the efficiency of token interaction.
 In \cref{tab:classification_cascade_parallel}, we test the effects between the cascade structure of CGSMM and the parallel structure of \textbf{P}arallel \textbf{G}roup \textbf{S}trip \textbf{M}ixing \textbf{M}odule (PGSMM).
 Our experiments show that the cascade structure gets a higher accuracy compared to the parallel structure, with an increase of +0.48\% and +0.61\% on the Caltech-101 and CIFAR-100 datasets, respectively.

 \begin{table}
   \centering
     \setlength{\tabcolsep}{2mm}{
     \renewcommand\arraystretch{0.4}
     \begin{tabular}{ c | c | c | c | c }
     \toprule
     Method & Dataset & Params & FLOPs & Top-1(\%) \\
     \midrule
     Parallel & \multirow{2}{*}{Caltech-101} & 42.67M & 6.65G & 91.61 \\
     Cascade & & 43.66M & 6.83G & 92.09 \\
     \midrule
     Parallel & \multirow{2}{*}{CIFAR-100} & 40.90M & 6.66G & 85.89 \\
     Cascade & & 41.88M & 6.83G & 86.50 \\
     \bottomrule
   \end{tabular}}
   \caption{Ablation study on the cascade and parallel structure of the Strip MLP in Group Strip MLP. 
   \emph{Parallel} means applying the Strip MLP on the row and column direction in parallel. \emph{Cascade} means the proposed CGSMM method in this paper. }
   \label{tab:classification_cascade_parallel}
     \vspace{-2ex}
 \end{table}
 
 \textbf{Roles of CGSMM \& LSMM.} To verify the significance of two complementary branches,
 we conduct experiments where we only keep one of the branches of CGSMM or LSMM.
 ~\cref{tab:local_mixing_module} displays the outcomes of our trial on Caltech-101 and CIFAR-100.
 Notably, we found that removing any module from either of the two branches significantly reduced the model's performance. These results highlight the essential role of both CGSMM and LSMM in enriching the tokens' interaction power.

 \begin{table}
  \centering
    \setlength{\tabcolsep}{0.4mm}{
    \renewcommand\arraystretch{0.4}
    \begin{tabular}{ c | c | c | c | c }
    \toprule
    Method & Dataset & Params & FLOPs & Top-1(\%) \\
    \midrule
    CGSMM Only & \multirow{3}{*}{Caltech-101} & 43.40M & 7.55G & 91.10 \\
    LSMM Only & & 43.96M & 6.00G & 90.31 \\
    CGSMM \& LSMM &  & 41.88M & 6.83G & 91.38 \\
    \midrule
    CGSMM Only & \multirow{3}{*}{CIFAR-100} & 43.40M & 7.55G & 85.20 \\
    LSMM Only & & 43.96M & 6.00G & 86.13 \\
    CGSMM \& LSMM &  & 41.88M & 6.83G & 86.50 \\
    \bottomrule
  \end{tabular}}
  \caption{Ablation study on the roles of two branches of CGSMM and LSMM. The base model is Strip-MLP-S. When any module branch has been removed, all input features would be fed into the remaining branch module.}
  \label{tab:local_mixing_module}
    \vspace{-3ex}
 \end{table}

 \section{Conclusion}
 \label{sec:Conclusion}
 
 This paper presents an efficient and effective token interaction MLP model of Strip-MLP for vision MLP. 
 For \emph{the Token’s interaction dilemma} problem, we design the Strip Mixing Block to enrich the token interaction power by three strategies.
 Our experimental analysis reveals that the Strip-MLP
 remarkably improves the performances of the MLP-based models on small datasets and keeps comparable or even better results on ImageNet with great superiorities on the number of parameters and FLOPs. 
 These observations clearly demonstrate that our models make token interaction more
 efficient among MLP-based models in image classification tasks.
 It is expected that Strip MLP layer has the potential to be a standard layer in the future and have promising performance on diverse vision tasks.
 
\textbf{Acknowledgement:} This work is supported by National Key Research and Development Program of China (2021YFF1200800) and Peng Cheng Laboratory Research Project (PCL2023AS6-1).

{\small
\bibliographystyle{ieee_fullname}
\bibliography{egbib}
}

\newpage
\appendix

\section{Pseudo-code of CGSMM}
\label{sec:pseudo_code}

In the main paper, we have introduced the \textbf{C}ascade \textbf{G}roup \textbf{S}trip \textbf{M}ixing \textbf{M}odule (CGSMM) in detail. Here, we further show the pseudo-code of the module in~\cref{alg:CGSMM} (the strip width and patch number are set to 3 and C/4, respectively). We will release the source code of the whole method upon acceptance.

\begin{algorithm*}
    \caption{Pseudo-code for CGSMM module.}\label{alg:cap}
    \begin{algorithmic}
    \Require x: the input tensor with shape (N, C, H, W) 
    \Ensure y: the output tensor with shape (N, C, H, W) 

    \State
    \Procedure {$CGSMM$}{$nn.Module$}:
      \State $def \_\_init\_\_(self, C, H, W):$
      \State \ \ \ \ $self.P = C / 4$   \ \ \ \ $\# patch \ number$
      \State \ \ \ \ $self.proj\_h = nn.Conv2d(H*self.P, H*self.P, (1,3), groups=self.P)$
      \State \ \ \ \ $self.proj\_w = nn.Conv2d(W*self.P, W*self.P, (1,3), groups=self.P)$
      \State \ \ \ \ $self.fuse\_h = nn.Conv2d(2C, C, (1,1))$
      \State \ \ \ \ $self.fuse\_w = nn.Conv2d(2C, C, (1,1))$
      \State
      \State $def forward(self, x):$
      \State \ \ \ \ $N, C, H, W = x.shape$
      \State \ \ \ \ $CP = C / self.P$
      \State \ \ \ \ $x = x.view(N, CP, self.P, H, W)$
      \State \ \ \ \ $x\_w = x.permute(0,1,3,2,4).view(N,CP, H,self.P*W)$
      \State \ \ \ \ $x\_w = self.proj\_w(x\_w.permute(0,3,1, 2)).permute(0,2,3,1)$
      \State \ \ \ \ $x\_w = x\_w.view(N,CP,H,self.P,W).permute(0,1,3, 2,4).view(N,C,H,W)$
      \State
      \State \ \ \ \ $x = x.view(N,C,H,W)$
      \State \ \ \ \ $x\_w = self.fuse\_w(torch.cat([x\_w, x], 1))$
      \State
      \State \ \ \ \ $x\_h = self.proj\_h(x\_w.view(N,CP,H*self.P,W).permute(0,2,1,3)).permute(0,2,1,3)$
      \State \ \ \ \ $x\_h = x\_h.view(N,CP,self.P,H, W).view(N,C,H,W)$

      \State \ \ \ \ $y = self.fuse\_h(torch.cat([x\_h, x], 1))$
      \State \ \ \ \ $return \ \ y$
      \EndProcedure    
    \end{algorithmic}
    \label{alg:CGSMM}
\end{algorithm*}

\section{Experiments}

\begin{figure*}
 \centering
	\subfigure[]{
		\begin{minipage}[t]{0.24\linewidth}
			\centering
			\includegraphics[width=1.0\linewidth]{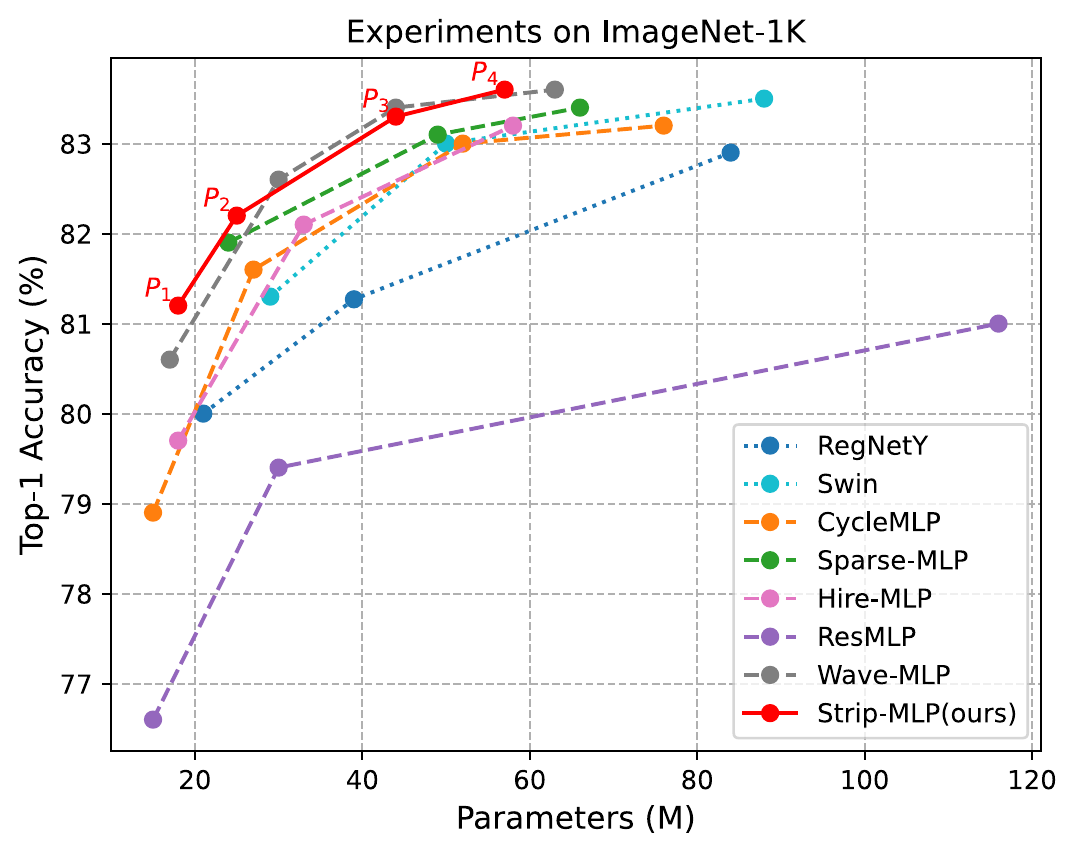}
		\end{minipage}
  \hspace{-2ex}
	}%
	\subfigure[]{
		\begin{minipage}[t]{0.24\linewidth}
			\centering
			\includegraphics[width=1.0\linewidth]{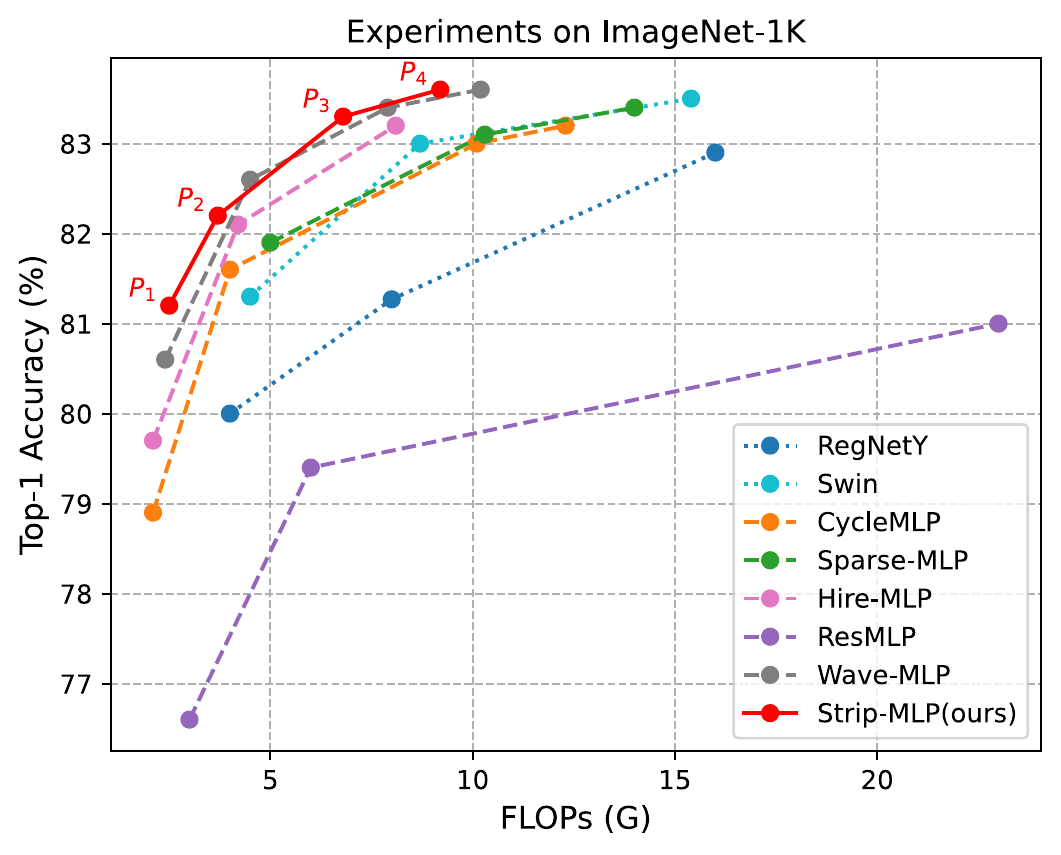}
		\end{minipage}
  \hspace{-2ex}
    }
  	\subfigure[]{
		\begin{minipage}[t]{0.24\linewidth}
			\includegraphics[width=1.0\linewidth]{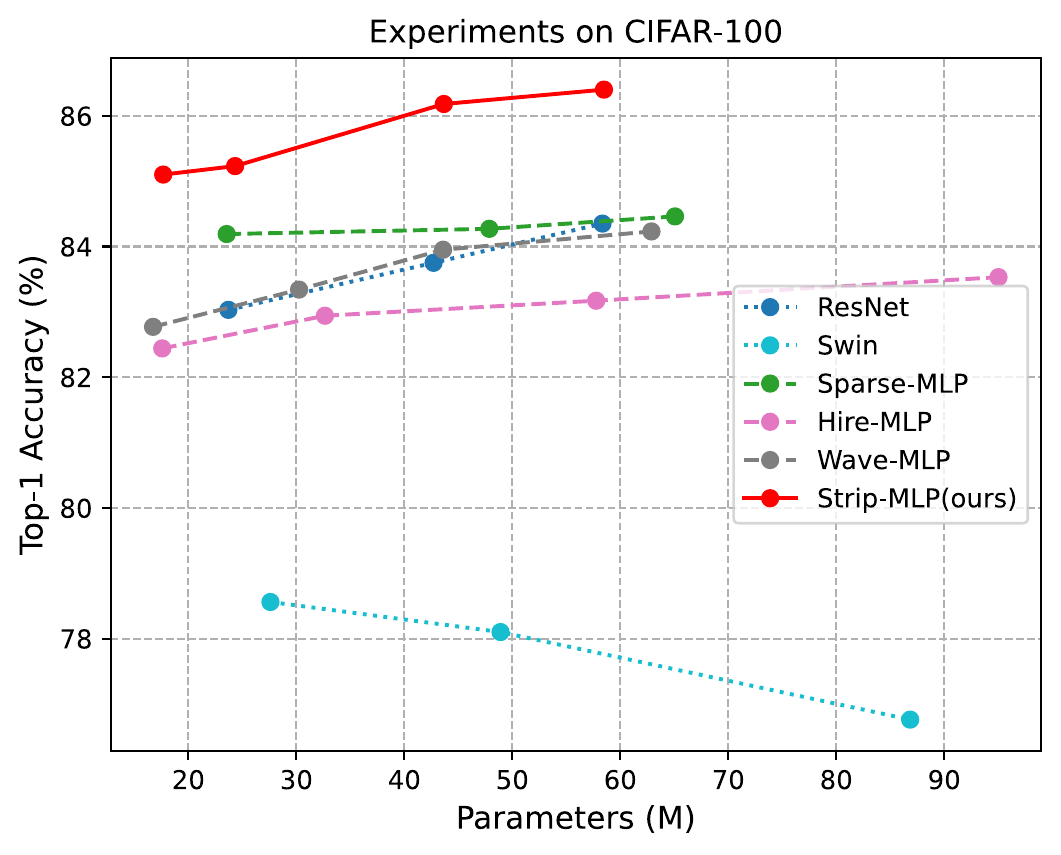}
		\end{minipage}
    \hspace{-2ex}
    }
  	\subfigure[]{
		\begin{minipage}[t]{0.24\linewidth}
			\includegraphics[width=1.0\linewidth]{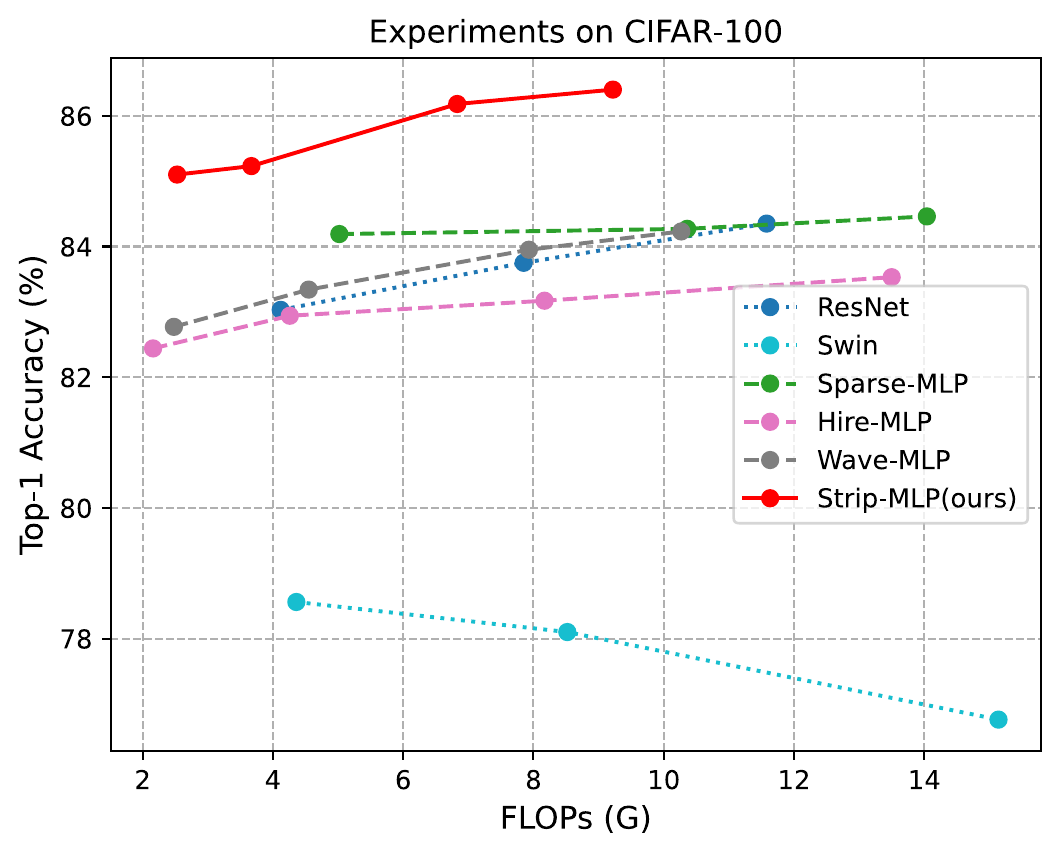}
		\end{minipage}    
	}

  	\subfigure[]{
		\begin{minipage}[t]{0.24\linewidth}
			\includegraphics[width=1.0\linewidth]{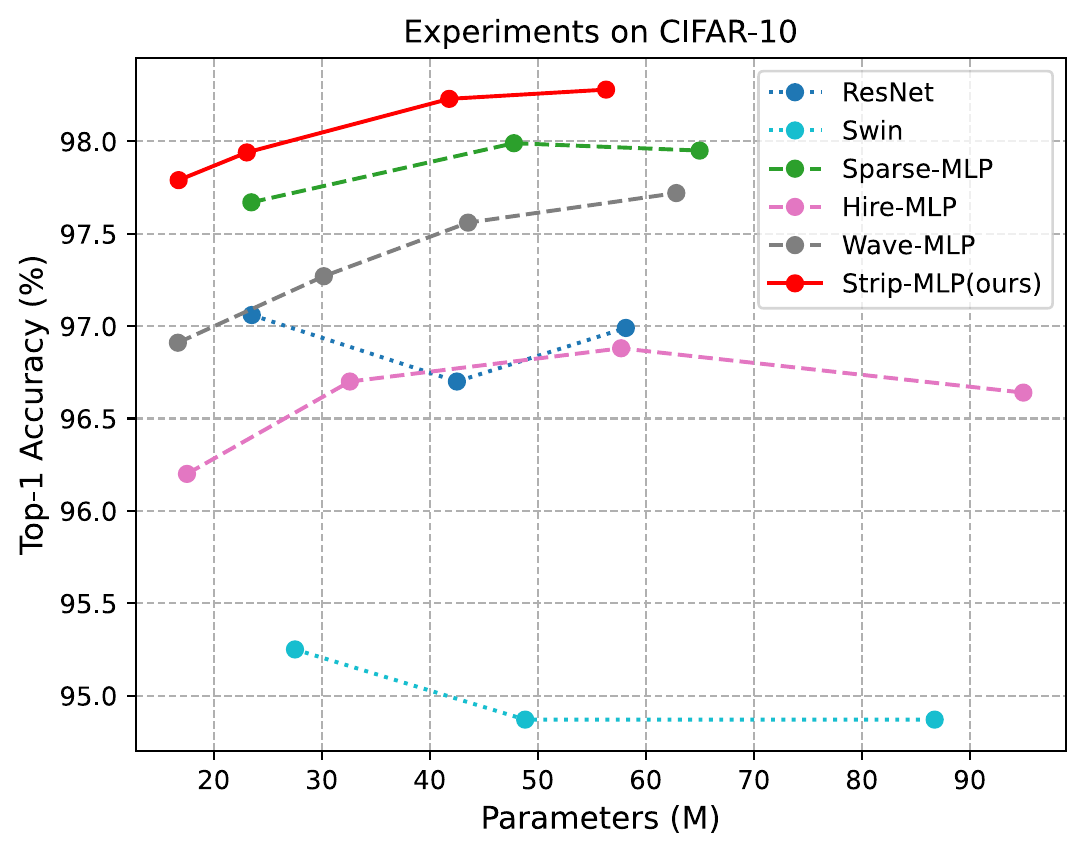}
		\end{minipage}
    \hspace{-2ex}
    }
  	\subfigure[]{
		\begin{minipage}[t]{0.24\linewidth}
			\includegraphics[width=1.0\linewidth]{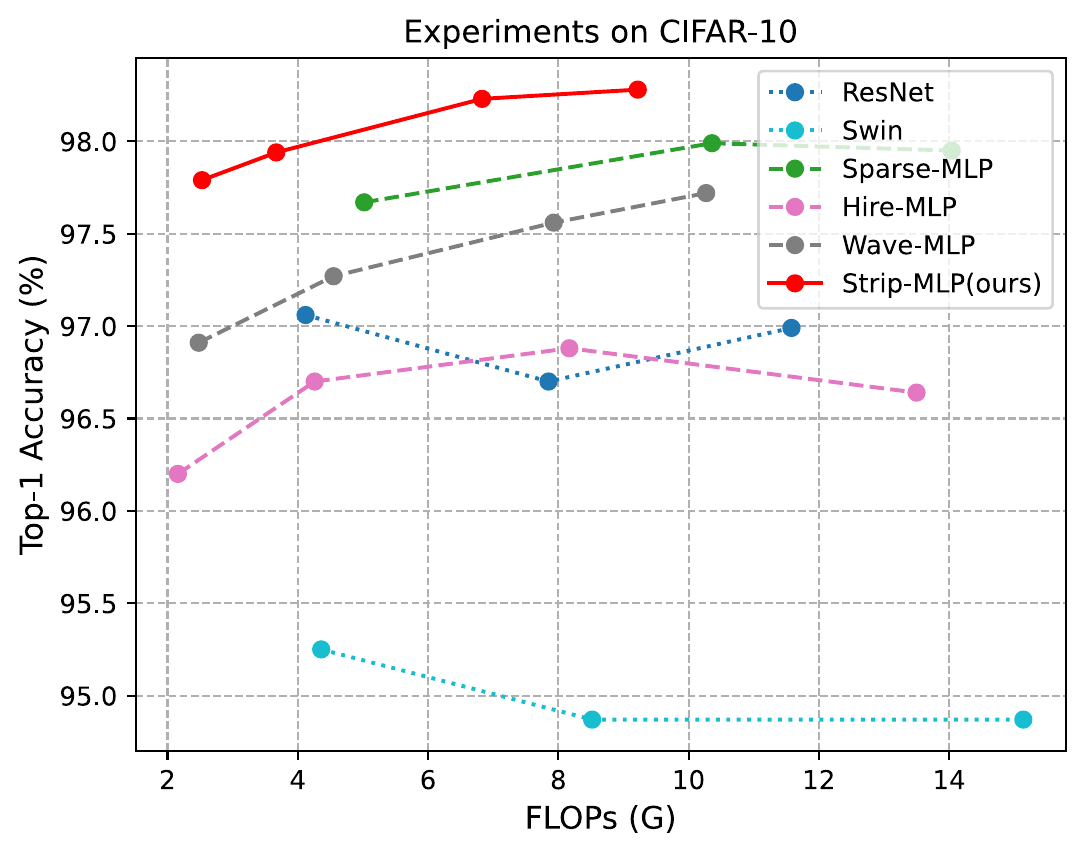}
		\end{minipage}    
    \hspace{-2ex}
	}
	\subfigure[]{
		\begin{minipage}[t]{0.24\linewidth}
			\centering
			\includegraphics[width=1.0\linewidth]{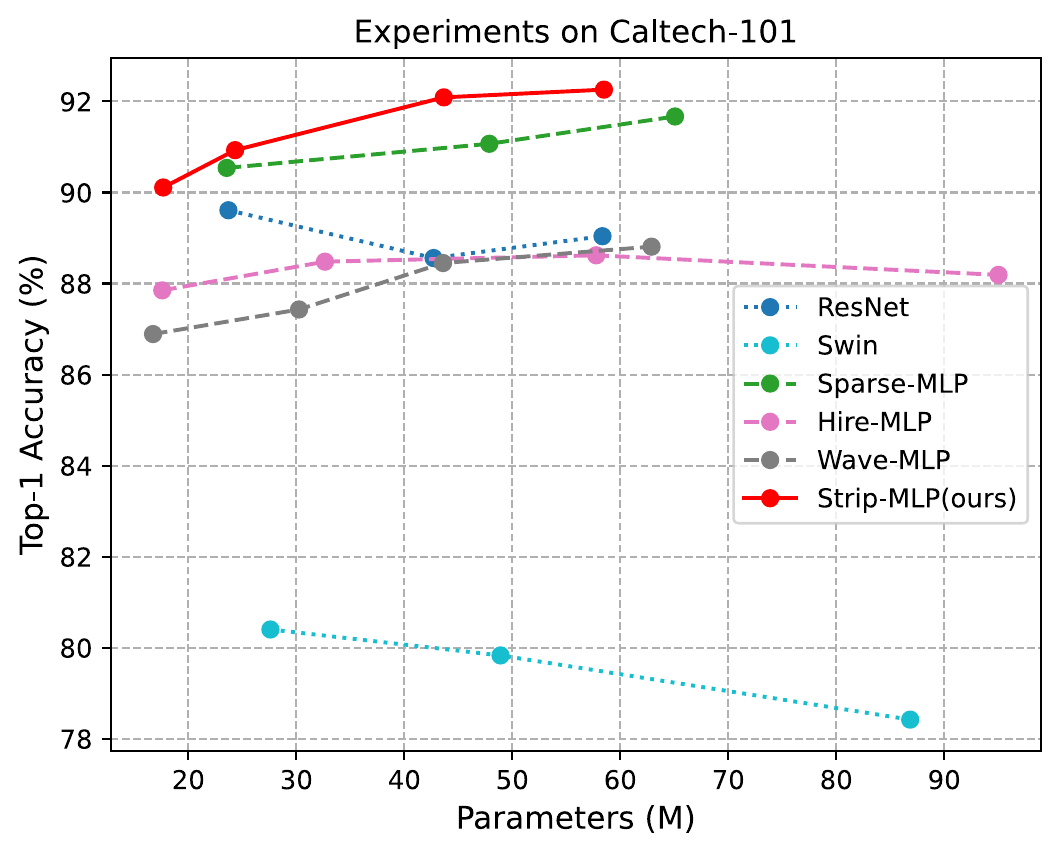}
		\end{minipage}
  \hspace{-2ex}
	}%
	\subfigure[]{
		\begin{minipage}[t]{0.24\linewidth}
			\centering
			\includegraphics[width=1.0\linewidth]{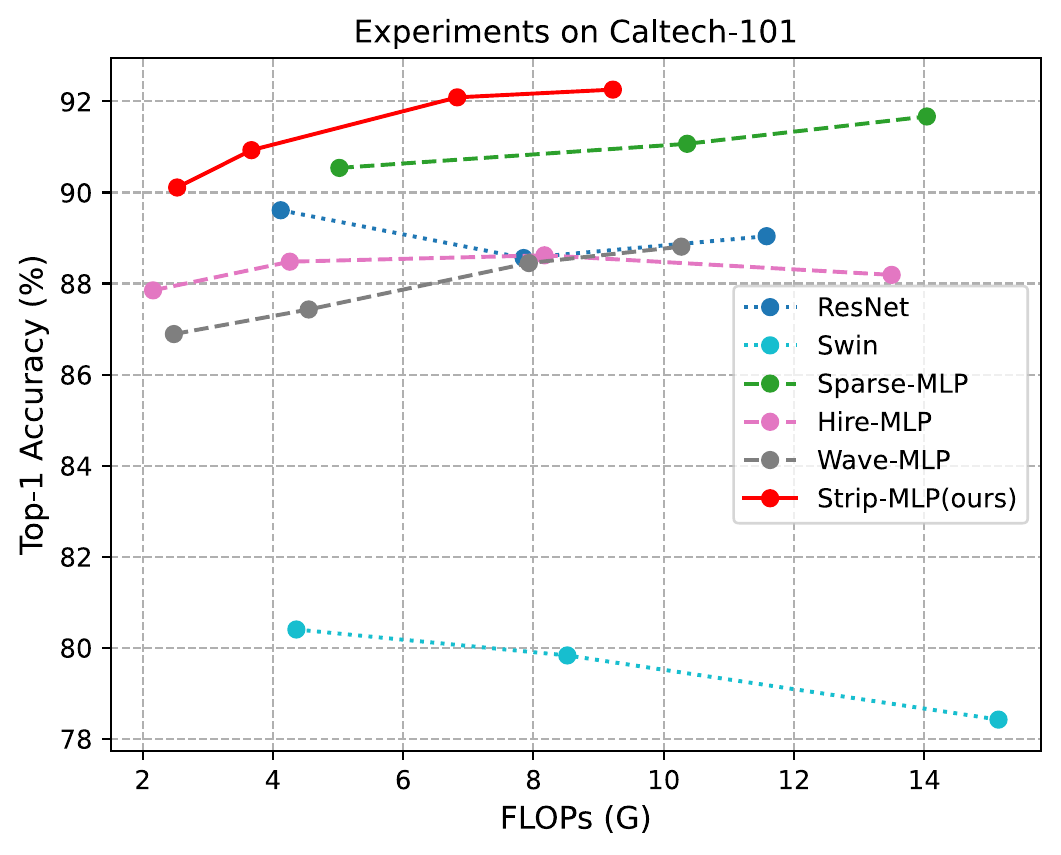}
		\end{minipage}
    }
     \caption{Performance comparison on four datasets of ImageNet-1K~\cite{deng2009imagenet}, CIFAR-100~\cite{krizhevsky2009learning}, CIFAR-10~\cite{krizhevsky2009learning}, and Caltech-101~\cite{fei2006one}. The Top-1 accuracy, number of parameters and FLOPs are reported with networks of CNN-based, Transformer-based, and MLP-based models.
      }
    \label{fig:performance}
\end{figure*}

In the main paper, we have described the experimental settings and reported results on three datasets with CNN-based, Transformer-based, and MLP-based models. As shown in \cref{fig:performance}, our Strip-MLP has significant superiority in performance and complexity over other state-of-the-art methods on different datasets, demonstrating our Strip-MLP is remarkably efficient and effective to enrich the token interaction power for vision MLP. In this section, we present the  experiments on the dataset of CIFAR-10.



 \begin{table}[h]
   \centering
     \renewcommand\arraystretch{0.5}
     \begin{tabular}{c | c | c | c }
     \toprule
     Method & Params & FLOPs & Top-1 (\%) \\
     \midrule
     \multicolumn{4}{c}{CNNs-Based} \\
     \midrule
     ResNet50~\cite{he2016deep} & 23.53M & 4.12G & 97.06 \\
     ResNet101~\cite{he2016deep} & 42.52M & 7.85G & 96.70 \\
     ResNet152~\cite{he2016deep} & 58.16M & 11.58G & 96.99 \\
     \midrule
     \multicolumn{4}{c}{Transformer-Based} \\
     \midrule
     Swin-T~\cite{liu2021swin} & 27.53M & 4.36G & 95.25 \\
     Swin-S~\cite{liu2021swin} & 48.84M & 8.52G & 94.87 \\
     Swin-B~\cite{liu2021swin} & 86.75M & 15.14G & 94.87 \\
     \midrule
     \multicolumn{4}{c}{MLP-Based} \\
     \midrule
     Wave-MLP-T~\cite{tang2022image} & 16.69M & 2.48G & 96.91 \\
     Hire-MLP-T~\cite{guo2022hire} & 17.53M & 2.16G & 96.20 \\
     Strip-MLP-T$^{*}$(\textbf{ours}) & 16.76M & 2.53G & \textbf{97.79} \\
     \midrule
     Wave-MLP-S~\cite{tang2022image} & 30.20M & 4.55G & 97.27 \\
     Hire-MLP-S~\cite{guo2022hire} & 32.61M & 4.26G & 96.70 \\
     Sparse-MLP-T~\cite{tang2022sparse} & 23.49M & 5.02G & 97.67 \\
     Strip-MLP-T(\textbf{ours}) & \textbf{23.07M} & \textbf{3.67G} & \textbf{97.94} \\
     \midrule
     Wave-MLP-M~\cite{tang2022image} & 43.55M & 7.93G & 97.56 \\
     Hire-MLP-B~\cite{guo2022hire} & 57.72M & 8.17G & 96.88 \\
     Sparse-MLP-S~\cite{tang2022sparse} & 47.80M & 10.36G & 97.99 \\
     Strip-MLP-S(\textbf{ours}) & \textbf{41.81M} & \textbf{6.83G} & \textbf{98.23} \\
     \midrule
     Wave-MLP-B~\cite{tang2022image} & 62.83M & 10.27G & 97.72 \\
     Hire-MLP-L~\cite{guo2022hire} & 94.96M & 13.50G & 96.64 \\
     Sparse-MLP-B~\cite{tang2022sparse} & 64.99M & 14.04G & 97.95 \\
     Strip-MLP-B(\textbf{ours}) & \textbf{56.33M} & \textbf{9.22G} & \textbf{98.28} \\
     \bottomrule
   \end{tabular}
   \caption{Image classification results of different models on CIFAR-10. The patch number of CGSMM is C/4.
   All models are trained from scratch without any extra data. Strip-MLP-T$^{*}$ is the model of light tiny version with fewer parameters and FLOPs than Strip-MLP-T.}
   \label{tab:classification_CIFAR10}
 \end{table}

\subsection{Experiments on CIFAR-10}

Limited by the space of the main paper, we only show the experimental results on the small datasets of Caltech-101 and CIFAR-100. Here, to further illustrate the effectiveness of our Strip-MLP in improving the token interaction power, we report the experimental results on another small dataset of CIFAR-10~\cite{krizhevsky2009learning}, which consists of $60k$ $32 \times 32$ images in $10$ classes, and there are $50k$ images for training and $10k$ images for testing.
~\cref{tab:classification_CIFAR10} shows the results of performance comparisons.
Compared with CNNs-based models, all four variants of our Strip-MLP models achieve higher Top-1 accuracy with fewer parameters and FLOPs, with an average increase of +1.14\% (98.06\% vs. 96.92\%). 
When compared with transformer-based models, our Strip-MLP achieves better performance by +3.06\% (98.06\% vs. 95.00\%) with fewer parameters (34.49M vs. 54.37M) and FLOPs (5.56G vs. 9.34G).
Compared with MLP-based models, all of our Strip-MLP models obtain the best Top-1 accuracy with fewer
parameters and FLOPs. For example, the average Top-1 accuracy of Strip-MLP is higher than Hire-MLP/Wave-MLP/Sparse-MLP by +1.45\%/0.69\%/0.19\% (98.06\% vs. 96.61\%/97.37\%/97.87\%) with fewer average parameters (34.49M vs. 50.71M/38.32M/45.43M) and fewer average FLOPs (5.56G vs. 9.36G/6.31G/9.81G)

In particular, with only 16.69M parameters and 2.48G FLOPs, our Strip-MLP-T$^*$ noticeably surpasses all variants of ResNet~\cite{he2016deep} and Swin-Transformer~\cite{liu2021swin} models, and Strip-MLP-T$^*$ outperforms the transfer learning models of ViT-S/16~\cite{dosovitskiy2020image}/TST-14~\cite{yuan2021tokens} by +0.69\%/+0.29\%, consistently demonstrating our Strip-MLP has significant superiorities in performance and complexity.

\subsection{Experiments on ImageNet-1K}

In Tab.~\textcolor{red}{5} of the main paper, we presented the experimental results with three main categories of networks.
Compared to CNN-based models, Strip-MLP achieves higher performance with fewer parameters and FLOPs. For example, Strip-MLP-S gets higher accuracy than RegNetY-16G~\cite{radosavovic2020designing} by +0.4\% (83.3\% vs. 82.9\%) with nearly half parameters (44M vs. 84M) and FLOPs (6.8G vs 16.0G) of RegNetY-16G. Similar results can be found when compared to Swin Transformer~\cite{liu2021swin}.
For example, Strip-MLP-S achieves slightly higher accuracy than Swin-S~\cite{liu2021swin}  (83.3\% vs. 83.0\%) but with fewer parameters (44M vs. 50M) and FLOPs (6.8G vs. 8.7G). 

Our Strip-MLP is more efficient in token interaction when compared with other MLP-based models.
With only 18M parameters and 2.5G FLOPs, Strip-MLP-T$^*$ achieves 81.2\% Top-1 accuracy, which is significantly higher than MLP-based models of Hire-MLP-Tiny~\cite{guo2022hire}/Wave-MLP-T~\cite{tang2022image} by +1.5\%/+0.6\% with a similar number of parameters and FLOPs. Strip-MLP-B achieves the same accuracy as Wave-MLP-B but uses fewer parameters (57M vs. 63M) and FLOPs (9.2G vs. 10.2G). 

In addition, Strip-MLP-T and Strip-MLP-S get higher performance than other popular MLP-based models~\cite{chen2021cyclemlp, guo2022hire, tang2022sparse, touvron2022resmlp} with fewer parameters and FLOPs except for Wave-MLP models. As shown in Tab.~\textcolor{red}{5}, the performance of our Strip-MLP-T (82.2\%) is lower than Wave-MLP-S (82.6\%). The main reason is the model configuration difference between the two models. 
In particular, Strip-MLP-T has fewer parameters (25M vs. 30M) and FLOPs (3.7G vs 4.5G) than Wave-MLP-S. The number of parameters increasing by +5M would have a large impact on the accuracy of the model. For instance, Strip-MLP-T is higher than Strip-MLP-T$^*$ by +1.0\% (82.2\% vs. 81.2\%) with the number of parameters increasing by +7M (25M vs. 18M), which means if we increase the number of parameters of Strip-MLP-T to 30M, the model accuracy will have an appreciable improvement.
In \cref{fig:performance} (a) and (b), our Strip-MLP (red solid line in the figure) shows obvious superiority on the point $P_1$ and $P_2$ than Wave-MLP~\cite{tang2022image} (gray dashed line in the figure), and the accuracy is slightly lower in $P_3$ (83.3\% vs. 83.4\%) but with fewer FLOPs (6.8G vs. 7.9G).

Although Wave-MLP-M shows higher accuracy at the point $P_3$ in~\cref{fig:performance} (a), it does not mean our Strip-MLP is invalid.
Design strategies of Wave-MLP and Strip-MLP are different. Wave-MLP focuses on aggregating tokens dynamically by wave function with two parts of amplitude and phase, which aims to model varying contents from different input images, while Strip-MLP focuses on improving the power of token interaction with Strip MLP layer to alleviate \emph{the Token’s interaction dilemma problem}. To further verify the effectiveness of Strip-MLP on Wave-MLP, we apply the Strip MLP Layer into Wave-MLP-Tiny (namely Wave-Strip-MLP-Tiny) on ImageNet-1K and get better results than the original model: 81.2\% vs 80.6\% (+0.6\%), which reveals that our Strip-MLP method is effective and easily extended to serve as a new unit for deep MLP variants.

\end{document}